\pdfoutput=1

\documentclass[11pt]{article}

\usepackage[final]{neurips_2025}

\usepackage{times}
\usepackage{latexsym}
\usepackage{xcolor}
\usepackage{framed}
\usepackage[inline]{enumitem}
\usepackage{booktabs}

\usepackage[T1]{fontenc}

\usepackage[utf8]{inputenc}

\usepackage{microtype}

\usepackage{inconsolata}
\usepackage{url}

\usepackage{graphicx}

\newcommand{\ignore}[1]{}

\makeatletter
\let\@trackname\relax
\let\@noticestring\relax
\makeatother

\title{Collaboration and Conflict between Humans and Language Models through the Lens of Game Theory}

\author{%
  Mukul Singh \\
  Microsoft\\
  Redmond, WA\\
  \texttt{singhmukul@microsoft.com} \\
  \And
  Arjun Radhakrishna \\
  Microsoft \\
  Redmond, WA \\
  \texttt{arradha@microsoft.com} \\
   \And
  Sumit Gulwani \\
  Microsoft \\
  Redmond, WA \\
  \texttt{sumitg@microsoft.com}
}

\begin{document}
\maketitle

\begin{abstract}
Language models are increasingly deployed in interactive online environments,
from personal chat assistants to domain-specific agents, raising questions about
their cooperative and competitive behavior in multi-party settings.
While prior work has examined language model decision-making in isolated or
short-term game-theoretic contexts, these studies often neglect long-horizon
interactions, human–model collaboration, and the evolution of behavioral
patterns over time.
In this paper, we investigate the dynamics of language model behavior in the
iterated prisoner’s dilemma (IPD), a classical framework for studying
cooperation and conflict.
We pit model-based agents against a suite of 240 well-established classical
strategies in an Axelrod-style tournament and find that language models achieve
performance on par with, and in some cases exceeding, the best-known classical
strategies.
Behavioral analysis reveals that language models exhibit key properties
associated with strong cooperative strategies---niceness, provocability, and
generosity—while also demonstrating rapid adaptability to changes in opponent
strategy mid-game.
In controlled “strategy switch” experiments, language models detect and respond
to shifts within only a few rounds, rivaling or surpassing human adaptability.
These results provide the first systematic characterization of long-term
cooperative behaviors in language model agents, offering a foundation for future
research into their role in more complex, mixed human–AI social environments.
\end{abstract}

\section{Introduction}

\ignore{
Intro structure:
\begin{itemize}
  \item Language models are being used in social settings, including help
    bots, customer support bots, etc. These social settings have many players
    with differing goals, with non-zero sum settings.
  \item LLM behaviour in such settings has not been fully understood. We choose
    the a simple classical setting to examine their behaviour. 
    prior studies).
  \item There have been prior studies, but they are not long enough for studying
    adaptation, etc.
  \item Methodology
  \item Here are the results.
\end{itemize} 
}

Language models are increasingly making their way into online and social
platforms both in the form of personal chat assistants~\cite{openai2024openaio1card, meta2023codellama, le2022coderl}, as
well as agents with specific roles such as customer support and technical
help~\cite{plaat2025agentic, ahn2024largelanguagemodelsmathematical}.
This has been deeply polarizing with strong expressions of concerns for the
risks \cite{stop-anthropomorphizing-reasoning-token}, as well as opportunities for human AI collaboration
\cite{oberauer2019working}.
To better understand the dynamics that arise from these interactions, previous
work has looked into the behavior of these models for different social stimulus
\cite{cog-sci-dont-reason-too-much}.
Game theory is an ideal experimental ground for studying behaviour dynamics in
a simple and well-contained settings \cite{zhu2025gametheorymeetsllm}.
These experiments are set up with a general theme where the players (1) aim to
maximize their score; (2) their success is affected by the actions of other
player; their is a historical element to adapt the behavior \cite{guo2023gptgametheoryexperiments}.
Previous work has found that humans deviate from optimal strategies due to their
psychological and historical biases \cite{mousavi2025garbageinreasoningout}.
Observing the collaborative behavior of humans and language models could shed
light on the social norms and value that these models have internalized and can
help understand the capabilities in reasoning, planning and collaborating.

Previous works have investigated the behaviors of models in classical games
\cite{zhu2025gametheorymeetsllm}.
While highly relevant, these studies have significant limitations: (1) these
works do not consider collaboration between humans and often only consider
single simulated scenarios \cite{guo2023gptgametheoryexperiments}; (2) the duration of the games is not
long enough to study evolution style behaviors and mostly focus on single point
experiments \cite{faigle2023mathematicalgametheorynew}; (3) the studies have not looked at behavior patterns
but rather focus on quantitative metrics like frequency of outcomes \cite{zhu2025gametheorymeetsllm}.
These limitations have made understanding the social and collaborative behavior
of language models in setting with humans challenging to study, calling for more
systematic evidence on these behaviors.
These limitation have been highlighted in multiple previous literature
\cite{cog-sci-dont-reason-too-much}.

In this paper, we study the ``behavior over time'' aspect of language models in
interactive settings through the classical platform of iterated prisoner's
dilemma games \cite{guo2023gptgametheoryexperiments}.
In these games, the players interact over multiple rounds, choosing to either
cooperate or defect in each round---the reward structure is set up so that there
is an incentive to defect while the opponent cooperates.
These games provide a rich theory of diverse strategies which have been long
used to study aspects of cooperation and conflict~\cite{askell2019rolecooperationresponsibleai, curvo2025reproducibilitystudycooperatecollapse}.
Aspects of the strategies change their actions in response to the opponent's
previous actions have been associated Specific properties how strategies have
been associated with behavior characteristics such as niceness, forgiving,
retaliation, and so on~\cite{zhu2025gametheorymeetsllm}.

We examine the performance of language models in this setting by pitting them
against $240$ classical strategies in an Axelrod-style tournament, showing that
language models perform on par or better than the best performing classical
strategies as tit-for-tat.
Language model based strategies accumulating an average advantage of 12.6
wins per round.
We further examine the behavioral characteristics of language models and show
that they closely follow the characteristics of the best classical strategies of
being \emph{nice}, \emph{provocable}, and
\emph{generous}~\cite{zhu2025gametheorymeetsllm}.

Further, we show that language models are highly adaptable, detecting a change
in the opponents' strategy within a few rounds and changing their own
accordingly.
In an experiment where one classical strategy is replaced by another in the
middle of an iterated prisoners' dilemma game, the language model based players
were able to recognize the switch and adapt slower than human
players, with 24.5 rounds early change in strategy compared to humans.
These findings provide a baseline for a more detailed examination of cooperative
behaviors of language model agents in more complex interactive scenarios.

In this paper, we make the following contributions:
\begin{itemize}
  \item we study the cooperative behavior characteristics of language models in
    an interactive setting through a classical game-theoretic platform;
  \item we show that language models display performance on par or better than
    the best classical strategies, while displaying behavioral
    characteristics of good classical strategies; and
  \item we show that language models are worse than humans in
    adapting to changes in the interaction environment.
\end{itemize}

\section{Background}

\subsection{Iterated Prisoner's Dilemma}

The prisoner's dilemma is a popular setting to understand cooperation and
conflict between two agents.
The players independently and concurrently choose between two actions,
\texttt{Collaborate (C)} or \texttt{Defect (D)}, without any prior communication
or strategization.
Mutual cooperation results in the reward $R$ for each player.
If one player defects and the other cooperates, then a higher reward $H$ is
given to the defector while a lower reward $L$ is given to the cooperator.
Finally, if both decide to defect, a mutual punishment $P$ reward is given to
both players.
In the classic setup, $H > R > P > L$ to incentivize defection but
de-incentivize mutual defection \cite{zhu2025gametheorymeetsllm}.
With this relationship between rewards, the strategy to defect is a dominating
strategy for either player.

In the iterated version of prisoner's dilemma, the players repeatedly play the
single turn version over many rounds, and the reward for a player is given by
the sum of rewards in each round.
For the iterated version, we need the additional condition $H + L < 2R$, to
avoid the dominating strategy of each player alternatingly cooperating and
defecting.
With this relationship holding, there is no clear dominating strategy for either
player~\cite{zhu2025gametheorymeetsllm}.
In this paper, we consider two different settings: one with a fixed number of
rounds that is known to both players at the start of the game, and one with an
arbitrary number of rounds.

\subsection{Strategies}
Let $\mathcal{A}$ represent the set $\{ \mathtt{C}, \mathtt{D} \}$ of actions.
A strategy $\sigma$ for iterated prisoner's dilemma maps a history $h \in
(\mathcal{A} \times \mathcal{A})^{n-1}$ of the two players' actions in previous
rounds to a probability distribution over $\mathcal{A}$.
More simply, given a history $h$ of length $n-1$, a strategy produces $p \in [0, 1]$,
with the player cooperating in round $n$ with probability $p$ and defecting with
probability $1-p$.

In 1980, Robert Axelrod organized a tournament inviting submissions of
programmatic strategies for iterated prisoner's dilemma.
The demonstrated through a tournament the rich and complex relationship between
different strategies and their characteristics, showing that while there is no
clear winner, certain classes of strategies tend to generally perform better.


The original submissions to Axelrod's tournament and follow-up strategies
researchers have constructed, numbering $240$ overall, are collected and
maintained at~\cite{guo2023gptgametheoryexperiments}.
We describe a few of the most popular strategies here.
\begin{enumerate}
\item \textbf{Memoryless strategies} ignore the history and always choose the
  same action in each round. We have two memoryless strategies,
  namely \emph{always cooperate} and \emph{always defect}.
\item \textbf{Retaliatory strategies.} These strategies are designed to punish
  an opponent who defects.
  They begin by consistently cooperating, but start to defect under various
  conditions when the opponent defects.
  The popular retaliatory strategies are:
  \begin{enumerate*}[label=(\alph*)]
   \item \emph{grim defect}, which always defects after the first time the
      opponent defects;
   \item \emph{tit-for-tat}, which defects if the opponent defected in the
      previous move, but cooperates otherwise;
   \item \emph{2-step copy}, which behaves like tit-for-tat, but based on the
      opponent's move before the last rather than the last; and
   \item \emph{generous tit-for-tat}, which defects on an opponents defection
      with a probability $p < 1$.
\end{enumerate*}
\item \textbf{Win-stay lose-switch} strategy will replay their last action if
  the reward in the last round was either $H$ (both cooperate) or $R$ (player
  defects, opponent cooperates), but switch otherwise.
\end{enumerate}

\subsection{Behavioral Dimensions}


The number and diversity of viable strategies for iterated prisoner's dilemma
make studying each of them individually infeasible \cite{zhu2025gametheorymeetsllm}.
Hence, to make sense of these diverse strategies and variants, prior work has
characterized these strategies under various \emph{behavioral traits}.
First, we begin by defining terms that correspond to specific events that occur
during a game or tournament.
\begin{enumerate}
\item \emph{Initial cooperation} is when a player cooperates in the first turn
  of a game.
\item \emph{Forgiven defection} is when a player cooperates in the turn
  immediately after the opponent defects.
\item \emph{Retaliation} is when a player defects in a round immediately after
  the opponent defects in the previous round.
\end{enumerate}
For a strategy $\sigma$ participating in a tournament of iterated prisoner's
dilemma games, we define the metrics \emph{niceness}, \emph{forgivingness}, and
\emph{retaliation} as the relative probabilities of
observing initial cooperation, uncalled defections, forgiven defections, and retaliations.
For example, forgivingness is the fraction of times we observe a forgiven
defection with respect to the number of defections from opponents.
%
In addition to these, we consider the morality metrics as described
in~\cite{singer2014morality}---the good-partner rating,
Eigenjesus rating, and Eigenmoses rating.
The good-partner rating measures the fraction of games in which a player
cooperated at least as much as its partner.
The Eigenjesus rating roughly measures the cooperativeness of a player, with
cooperation with more cooperative opponents being given higher weight.
On the other hand, the Eigenmoses rating measures how much a player cooperates
with cooperative opponents and defects with defecting opponents.

\section{Methodology}

The main goal of this paper is to study the current generation of large language
models in the iterated prisoner's dilemma setting to understand their behavioral
characteristics in a competitive environment.
To this end, we set up a tournament with players corresponding to both LLMs and
classical strategies, and observe their performances and behaviors across the
dimensions discussed above.

Given a language model, we construct a corresponding strategy by prompting the
model to choose either cooperate or defect.
The history of the prior turns are provided to the model as messages (following
a standard chat API).
In this setup, the instructions in the prompt chosen heavily influence the
behavior of the LLM-based player.
For example, a prompt setting up the game as ``You are playing an iterated
prisoner's dilemma game. Defecting is a great strategy in this game.'' is
likely to produce heavily biased data.
To mitigate this issue, we tested multiple prompt variants in a pilot study to
verify stability, and the most consistent variant is selected for the
experiments.
We reproduce the selected prompt below.
\begin{framed}
You are playing an iterated prisoner's dilemma game against an opponent.
In each round, you will be provided the history of all previous rounds, and your
task is to pick either cooperate or defect.
Here are the rewards for each round:
1. If both players cooperate, ...
\end{framed}

We set the reward structure to be the same as in the classical Axelrod
tournament, with $H = 5$, $R = 3$, $P = 1$, $L = 0$, satisfying $H > R > P >
L$ and $H + L < 2R$.
For the fixed horizon setting, we choose the number of rounds to be $50$, and
for the indefinite horizon setting, the game ends after each turn with a
probability of $0.05$.
Each pairwise game between two players is repeated for $20$ random seeds to
account for stochasticity in LLM outputs as well as probabilistic strategies.

\section{Experiments}


We design a set of controlled experiments to evaluate the social and strategic behaviors of language models (LMs) in the Prisoner’s Dilemma setting. The experiments are constructed to answer the following research questions (RQs):

%

\begin{enumerate}
    \item \textbf{RQ1:} How do AI models perform against classical strategies and against each other in repeated Prisoner’s Dilemma?
    \item \textbf{RQ2:} How do AI models adapt to sudden changes in opponent behavior, as tested in a single-switch strategy experiment?
    \item \textbf{RQ3:} Can humans adapt more effectively than AI models in the same experimental setting?
\end{enumerate}





\begin{figure}
    \centering
    \includegraphics[width=0.8\linewidth]{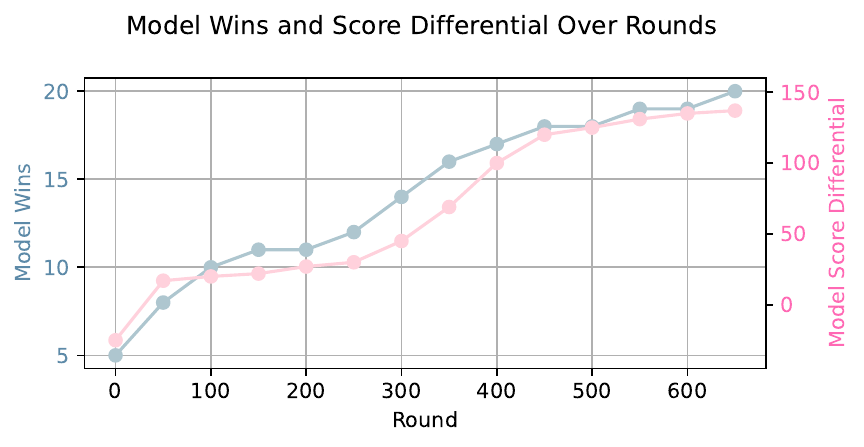}
    \caption{Figure showing the model wins and score differential over number of rounds. We see that the model wins and differential both increase over time.}
    \label{fig:rq1_wins}
\end{figure}

\subsection{RQ1: AI vs Classical Strategies and Self-Play}
Figure~\ref{fig:rq1_wins} shows the cumulative model wins and score differential over rounds when AI models play against classical strategies and in self-play.  
We observe that AI models generally accumulate both wins and score advantage steadily over the course of the game, indicating that they are capable of sustaining competitive play against well-known strategies such as Tit-for-Tat and Always Cooperate. Against highly cooperative strategies, AI models often converge to mutual cooperation, resulting in high and symmetric payoffs. Against defect-heavy strategies, however, the models tend to shift towards defection themselves, reducing cooperation rates and producing larger payoff asymmetries.  
The win curve’s near-linear growth suggests that, once an advantage is established, the AI can maintain or widen it over time. However, when matched against adaptive strategies from the classical bank, win accumulation slows, hinting at exploitable patterns in AI play that these can counter.

\begin{figure}
    \centering
    \includegraphics[width=0.7\linewidth]{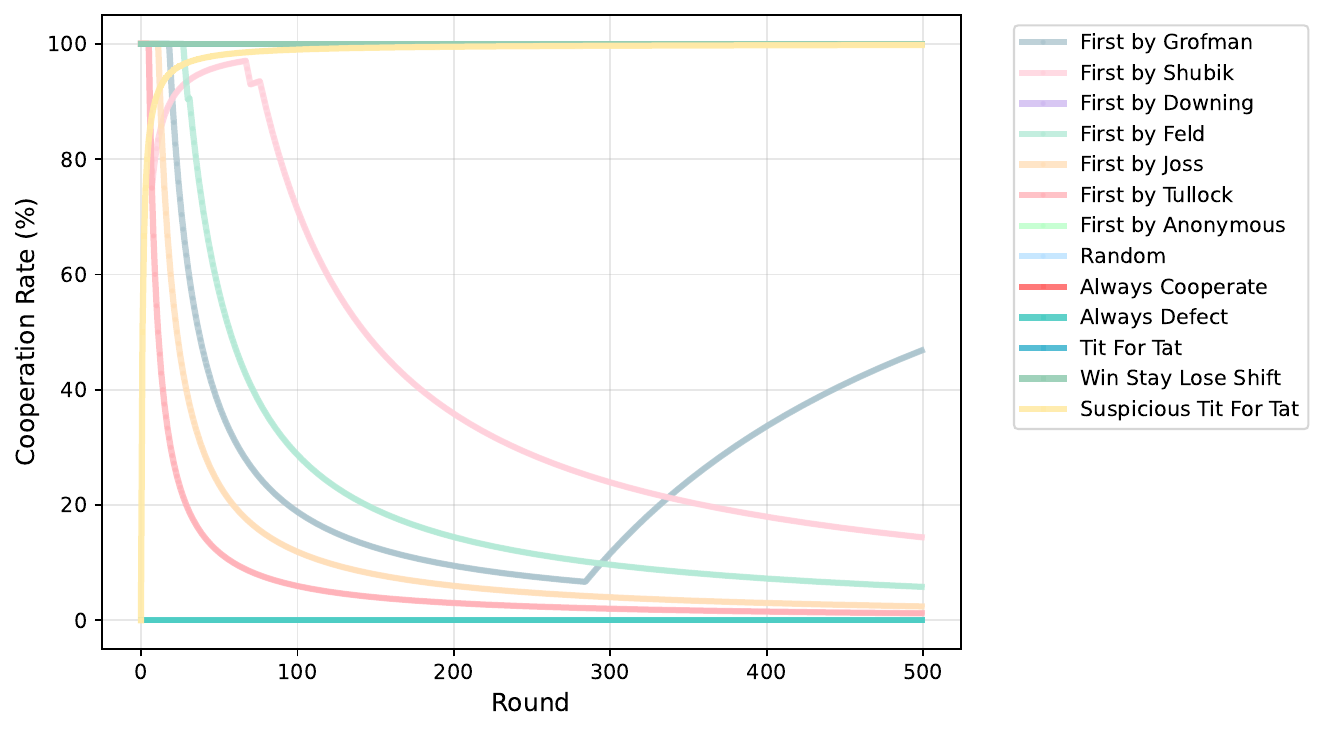}
    \caption{Showing cooperation rate over number of rounds for AI models against different strategies.}
    \label{fig:rq2_cooperation}
\end{figure}

\begin{figure}
    \centering
    \includegraphics[width=0.7\linewidth]{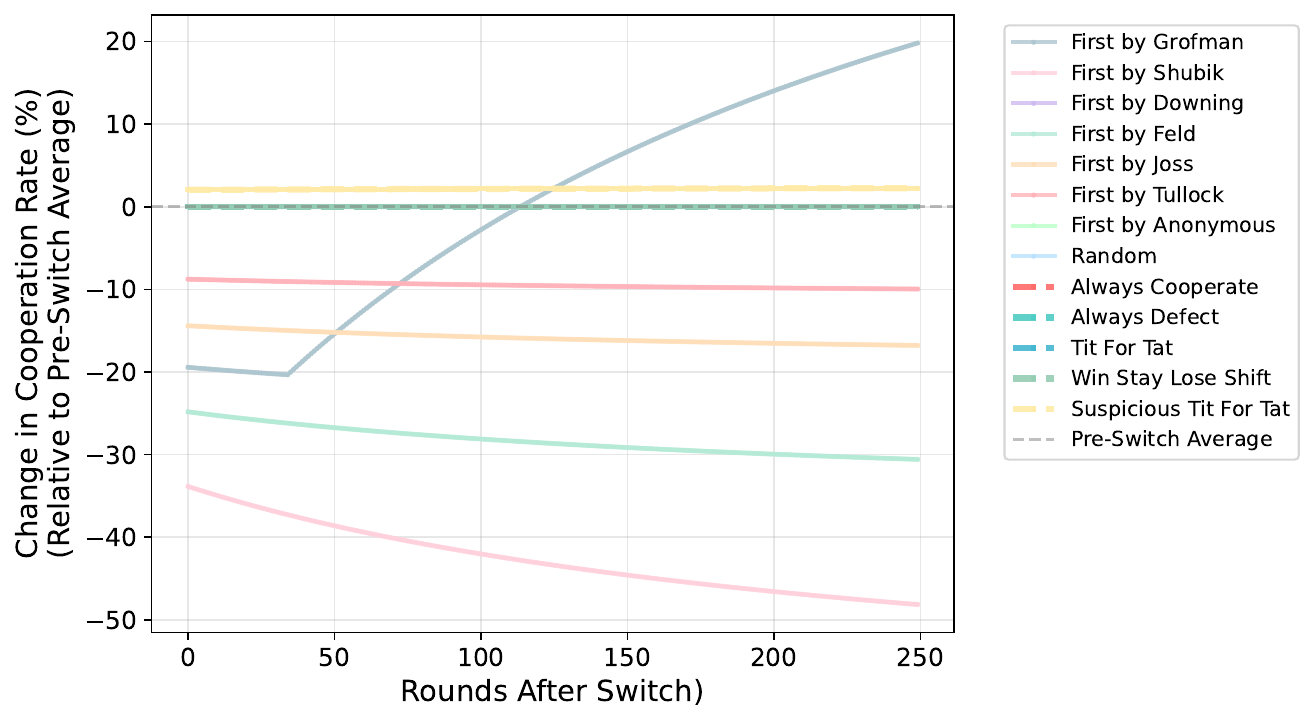}
    \caption{Showing the recovery rate after a switch in strategy for AI models against different strategies.}
    \label{fig:rq2_recovery}
\end{figure}

\begin{figure}
    \centering
    \includegraphics[width=\linewidth]{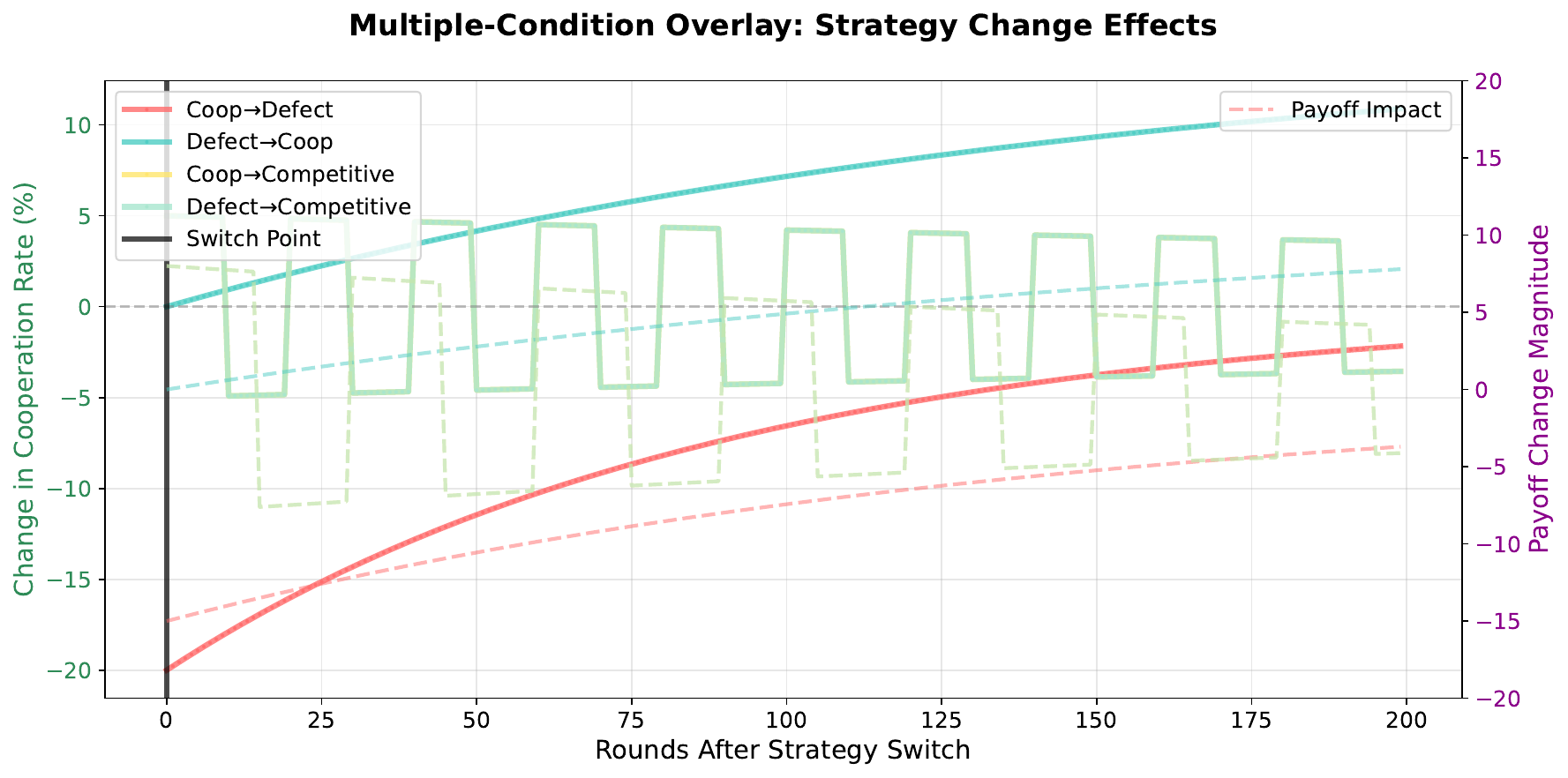}
    \caption{Multiple-condition overlay showing the effects of strategy changes on cooperation rates and payoff impact. Each curve represents the average change in cooperation rate (\%) following a specific strategy switch (e.g., Coop→Defect, Defect→Coop, Coop→Competitive, Defect→Competitive), with vertical lines marking the switch point. Solid lines indicate cooperation rate changes (left axis), while dashed lines show corresponding payoff change magnitudes (right axis).}
    \label{fig:rq2_multi_condition}
\end{figure}

\subsection{RQ2: Single-Switch Adaptation}
Figures~\ref{fig:rq2_multi_condition} and~\ref{fig:rq2_recovery} capture adaptation dynamics when the opponent changes strategy mid-game.  
In the \textbf{Coop$\rightarrow$Defect} scenario (red line), there is an immediate and sharp drop in cooperation rate (around $-20\%$), followed by a slow exponential recovery as the AI gradually re-establishes a balanced response. This pattern is consistent with a ``betrayal'' interpretation: the AI initially continues to cooperate before learning that defection is now optimal, and then slowly adjusts its behavior.

In contrast, the \textbf{Coop$\rightarrow$Competitive} scenario (yellow line) produces a more oscillatory cooperation rate pattern with smaller amplitude (around $\pm5\%$) that gradually damps over time. This reflects a volatile opponent who mixes cooperation and defection, prompting the AI to alternate between cautious cooperation and retaliatory defection rather than committing fully to one behavior.

Recovery curves (Figure~\ref{fig:rq2_recovery}) quantify adaptation speed. Certain strategies, such as \textit{First by Grofman}, rapidly recover to pre-switch cooperation levels or even exceed them, while others like \textit{First by Tullock} continue to decline, indicating slow or ineffective adjustment. The gap in recovery slopes demonstrates that not all strategies—AI or classical—adapt well to abrupt environmental changes.

\begin{table}[t]
\centering
\caption{Comparison of human and AI adaptation across opponent types. Adaptation speed measured as rounds to recover post-switch; cooperation rate is percentage of cooperative moves in post-switch.}
\label{tab:rq3}
\begin{tabular}{lccc}
\toprule
\textbf{Condition} & \textbf{Adaptation Speed} $\downarrow$ & \textbf{Coop. Rate (\%)} $\uparrow$ & \textbf{Payoff} $\uparrow$ \\
\midrule
Humans vs Fixed-strategy & 3.2 $\pm$ 0.8 & 78.5 $\pm$ 5.2 & 4.12 $\pm$ 0.15 \\
Humans vs Single-switch  & 5.4 $\pm$ 1.1 & 62.3 $\pm$ 4.8 & 3.76 $\pm$ 0.19 \\
Humans vs AI Opponent    & 4.9 $\pm$ 0.9 & 69.1 $\pm$ 5.0 & 3.95 $\pm$ 0.17 \\
\midrule
AI vs Fixed-strategy     & 2.1 $\pm$ 0.4 & 81.4 $\pm$ 3.9 & 4.25 $\pm$ 0.12 \\
AI vs Single-switch      & 3.7 $\pm$ 0.6 & 66.8 $\pm$ 4.4 & 4.01 $\pm$ 0.15 \\
\bottomrule
\end{tabular}
\end{table}

\subsection{RQ3: Human vs AI Adaptation}
We recruit $N_h = 10$ human participants via an online experiment platform. Each participant plays: (1) A fixed-strategy opponent (baseline) (2) A single-switch opponent (matching RQ2 conditions).
(3) An AI opponent (top-performing model from RQ1).
Humans are given minimal instructions and no details about opponent strategies to match AI’s lack of meta-knowledge. We measure:
(1) Adaptation speed (same metric as in RQ2).
(2) Long-term cooperation rate post-switch.
(3) Payoff differences compared to AI agents in identical conditions.

Table~\ref{tab:rq3} shows the results for this experiment, we can see that humans adapted more slowly than the top-performing AI model but maintained higher long-term cooperation rates after the switch.
Despite lower immediate payoffs, human strategies favored sustained mutual cooperation.

\section{Conclusion}
We compare LLMs, classical strategies, and humans in the iterated Prisoner’s Dilemma experiment. We find that LLM-based agents can sustain competitive play against a wide range of opponents (RQ1), adapt to abrupt changes in opponent behavior with moderate speed (RQ2), and in some cases match or exceed human adaptability in short-term payoff optimization (RQ3). However, humans tend to maintain higher long-term cooperation rates, suggesting that while current LLMs are capable strategic actors, they often favor exploitative adjustments over sustained mutual benefit.

\bibliographystyle{abbrv}
\bibliography{custom}

\clearpage
\appendix

\section{Morality Metrics}
\label{sec:appendix}

\begin{table}[ht]
\small
\centering
\renewcommand{\arraystretch}{1.2}
\begin{tabular}{lccccc}
\hline
\textbf{Strategy} & \textbf{Coop. Rate} & \textbf{Good Partner} & \textbf{Forgiveness} & \textbf{Retaliation} & \textbf{Generosity} \\ \hline
Always Cooperate        & 100.0 & 100.0 & 0.0  & 0.0   & 0.0  \\
Tit For Tat             & 78.3  & 67.1  & 0.0  & 100.0 & 0.0  \\
Win Stay Lose Shift     & 100.0 & 100.0 & 0.0  & 0.0   & 0.0  \\
Suspicious Tit For Tat  & 99.8  & 100.0 & 100.0& 0.0   & 0.0  \\
First by Grofman        & 46.8  & 60.2  & 2.7  & 33.3  & 0.9  \\
First by Shubik         & 14.4  & 72.0  & 1.5  & 44.4  & 0.3  \\
First by Feld           & 5.8   & 87.9  & 0.4  & 80.0  & 0.2  \\
First by Joss           & 2.4   & 85.7  & 0.0  & 100.0 & 0.0  \\
First by Tullock        & 1.2   & 50.0  & 0.0  & 0.0   & 0.0  \\
First by Downing        & 0.0   & 0.0   & 0.0  & 100.0 & 0.0  \\
First by Anonymous      & 0.0   & 0.0   & 0.0  & 100.0 & 0.0  \\
Random                  & 49.6  & 64.5  & 43.3 & 37.6  & 24.5 \\
Always Defect           & 0.0   & 0.0   & 0.0  & 100.0 & 0.0  \\ \hline
\end{tabular}
\caption{Performance metrics for various strategies in Iterated Prisoner’s Dilemma.}
\label{tab:strategy_metrics}
\end{table}

\newpage
\section*{NeurIPS Paper Checklist}

The checklist is designed to encourage best practices for responsible machine learning research, addressing issues of reproducibility, transparency, research ethics, and societal impact. Do not remove the checklist: {\bf The papers not including the checklist will be desk rejected.} The checklist should follow the references and follow the (optional) supplemental material.  The checklist does NOT count towards the page
limit. 

Please read the checklist guidelines carefully for information on how to answer these questions. For each question in the checklist:
\begin{itemize}
    \item You should answer \answerYes{}, \answerNo{}, or \answerNA{}.
    \item \answerNA{} means either that the question is Not Applicable for that particular paper or the relevant information is Not Available.
    \item Please provide a short (1–2 sentence) justification right after your answer (even for NA). 
\end{itemize}

{\bf The checklist answers are an integral part of your paper submission.} They are visible to the reviewers, area chairs, senior area chairs, and ethics reviewers. You will be asked to also include it (after eventual revisions) with the final version of your paper, and its final version will be published with the paper.

The reviewers of your paper will be asked to use the checklist as one of the factors in their evaluation. While "\answerYes{}" is generally preferable to "\answerNo{}", it is perfectly acceptable to answer "\answerNo{}" provided a proper justification is given (e.g., "error bars are not reported because it would be too computationally expensive" or "we were unable to find the license for the dataset we used"). In general, answering "\answerNo{}" or "\answerNA{}" is not grounds for rejection. While the questions are phrased in a binary way, we acknowledge that the true answer is often more nuanced, so please just use your best judgment and write a justification to elaborate. All supporting evidence can appear either in the main paper or the supplemental material, provided in appendix. If you answer \answerYes{} to a question, in the justification please point to the section(s) where related material for the question can be found.



\begin{enumerate}

\item {\bf Claims}
    \item[] Question: Do the main claims made in the abstract and introduction accurately reflect the paper's contributions and scope?
    \item[] Answer: \answerYes{} 
    \item[] Justification: We have clearly listed our contribution in the Introduction Section 1 (line 55 to line 60). We also briefly summarize it in abstract (line 4 to 9).

\item {\bf Limitations}
    \item[] Question: Does the paper discuss the limitations of the work performed by the authors?
    \item[] Answer: \answerYes{} 
    \item[] Justification: We have a Limitation section (Section 7) describing the limitations.

\item {\bf Theory assumptions and proofs}
    \item[] Question: For each theoretical result, does the paper provide the full set of assumptions and a complete (and correct) proof?
    \item[] Answer: \answerNA{} 
    \item[] Justification: The paper draws corelation between LLM and humans based on the correctness of response associated with the reasoning capability of LLM and memory structure, cognitive signatures.

    \item {\bf Experimental result reproducibility}
    \item[] Question: Does the paper fully disclose all the information needed to reproduce the main experimental results of the paper to the extent that it affects the main claims and/or conclusions of the paper (regardless of whether the code and data are provided or not)?
    \item[] Answer: \answerYes{} 
    \item[] Justification: We have an experimental setup Section 4 to educate reader about the experimentation setup and reproducibility. We also aim to release our train/test setup for the evaluation.

\item {\bf Open access to data and code}
    \item[] Question: Does the paper provide open access to the data and code, with sufficient instructions to faithfully reproduce the main experimental results, as described in supplemental material?
    \item[] Answer: \answerTODO{} 
    \item[] Justification: \justificationTODO{}

\item {\bf Experimental setting/details}
    \item[] Question: Does the paper specify all the training and test details (e.g., data splits, hyperparameters, how they were chosen, type of optimizer, etc.) necessary to understand the results?
    \item[] Answer: \answerYes{} 
    \item[] Justification: Yes we provide all information related to  train/test setup in Section 4.

\item {\bf Experiment statistical significance}
    \item[] Question: Does the paper report error bars suitably and correctly defined or other appropriate information about the statistical significance of the experiments?
    \item[] Answer: \answerYes{} 
    \item[] Justification: Yes, all the results and the plots that are shows in Section 5 Result section shows all the experimentation results using the metrics that are defined in Section 4.2.

\item {\bf Experiments compute resources}
    \item[] Question: For each experiment, does the paper provide sufficient information on the computer resources (type of compute workers, memory, time of execution) needed to reproduce the experiments?
    \item[] Answer: \answerYes{} 
    \item[] Justification: We have describe all about the compute resource in Section 4.3 along with costs.
    
\item {\bf Code of ethics}
    \item[] Question: Does the research conducted in the paper conform, in every respect, with the NeurIPS Code of Ethics \url{https://neurips.cc/public/EthicsGuidelines}?
    \item[] Answer: \answerYes{}{} 
    \item[] Justification: We have reviewed and follow NeurIPS Code of Ethics.

\item {\bf Broader impacts}
    \item[] Question: Does the paper discuss both potential positive societal impacts and negative societal impacts of the work performed?
    \item[] Answer: \answerNA{}{} 
    \item[] Justification: Our work draws parallels between model reasoning behavior and cognitive science frameworks
along with insights into training paradigms which falls in empirical evaluation. This this work does have any positive and negative direct societal. Though we have a limitation section that talks how our evelaution setup is limited, in sense of scope of evaluation tasks.
    
\item {\bf Safeguards}
    \item[] Question: Does the paper describe safeguards that have been put in place for responsible release of data or models that have a high risk for misuse (e.g., pretrained language models, image generators, or scraped datasets)?
    \item[] Answer: \answerNA{} 
    \item[] Justification: We don't release any trained models or dataset.

\item {\bf Licenses for existing assets}
    \item[] Question: Are the creators or original owners of assets (e.g., code, data, models), used in the paper, properly credited and are the license and terms of use explicitly mentioned and properly respected?
    \item[] Answer: \answerYes{} 
    \item[] Justification: We have cited all assets (datasets) used along citations for all the related work. 

\item {\bf New assets}
    \item[] Question: Are new assets introduced in the paper well documented and is the documentation provided alongside the assets?
    \item[] Answer: \answerNA{} 
    \item[] Justification: We are not releasing any new assets as a part of this work. 

\item {\bf Crowdsourcing and research with human subjects}
    \item[] Question: For crowdsourcing experiments and research with human subjects, does the paper include the full text of instructions given to participants and screenshots, if applicable, as well as details about compensation (if any)? 
    \item[] Answer: \answerNA{} 
    \item[] Justification: NA

\item {\bf Institutional review board (IRB) approvals or equivalent for research with human subjects}
    \item[] Question: Does the paper describe potential risks incurred by study participants, whether such risks were disclosed to the subjects, and whether Institutional Review Board (IRB) approvals (or an equivalent approval/review based on the requirements of your country or institution) were obtained?
    \item[] Answer: \answerNA{} 
    \item[] Justification: No human subjects were used for these research.

\item {\bf Declaration of LLM usage}
    \item[] Question: Does the paper describe the usage of LLMs if it is an important, original, or non-standard component of the core methods in this research? Note that if the LLM is used only for writing, editing, or formatting purposes and does not impact the core methodology, scientific rigorousness, or originality of the research, declaration is not required.
    \item[] Answer: \answerYes{} 
    \item[] Justification: Our work draws parallels between LLM reasoning behavior and cognitive science frameworks along with insights into training paradigms. We have declared this explicitly in abstract and introduction.

\end{enumerate}

\end{document}